\documentclass[conference]{IEEEtran}
\IEEEoverridecommandlockouts
\usepackage{amsmath,amssymb,amsfonts}
\usepackage{algorithmic}
\usepackage{graphicx}
\usepackage{textcomp}
\usepackage{xcolor}
\usepackage{array}
\usepackage[caption=false,font=normalsize,labelfont=sf,textfont=sf]{subfig}
\usepackage{booktabs}
\usepackage[square,sort,comma,numbers]{natbib}
\usepackage{multirow}
\usepackage{hyperref}
\usepackage{adjustbox}
\usepackage{bigstrut} 

\def\BibTeX{{\rm B\kern-.05em{\sc i\kern-.025em b}\kern-.08em
    T\kern-.1667em\lower.7ex\hbox{E}\kern-.125emX}}
\begin{document}

\title{Learning Personalized Utility Functions for Drivers in Ride-hailing Systems Using Ensemble Hypernetworks\\
}


\author{Weiming Mai, Jie Gao, Oded Cats

\thanks{Weiming Mai, Jie Gao, Oded Cats are with the Department of Transportation and Planning, Delft University of Technology, Delft, Netherlands (e-mail: w.m.mai@tudelft.nl; j.gao-1@tudelft.nl;  o.cats@tudelft.nl).}}
\maketitle

\begin{abstract}
In ride-hailing systems, drivers decide whether to accept or reject ride requests based on factors such as order characteristics, traffic conditions, and personal preferences. Accurately predicting these decisions is essential for improving the efficiency and reliability of these systems. Traditional models, such as the Random Utility Maximization (RUM) approach, typically predict drivers' decisions by assuming linear correlations among attributes. However, these models often fall short because they fail to account for non-linear interactions between attributes and do not cater to the unique, personalized preferences of individual drivers. In this paper, we develop a method for learning personalized utility functions using hypernetwork and ensemble learning. Hypernetworks dynamically generate weights for a linear utility function based on trip request data and driver profiles, capturing the non-linear relationships. An ensemble of hypernetworks trained on different
data segments further improve model adaptability and generalization by introducing controlled
randomness, thereby reducing over-fitting.


We validate the performance of our ensemble hypernetworks model in terms of prediction accuracy and uncertainty estimation in a real-world dataset. The results demonstrate that our approach not only accurately predicts each driver’s utility but also effectively balances the needs for explainability and uncertainty quantification. Additionally, our model serves as a powerful tool for revealing the personalized preferences of different drivers, clearly illustrating which attributes largely impact their rider acceptance decisions.

\end{abstract}


\section{Introduction}

Making a decision to accept or decline a request involves a binary choice. Typically, to identify the underlying determinants of the ride acceptance behaviour of drivers, the discrete choice theory is used to model the utility of the driver by a linear weighted combination of the value of each of the determinate attributes. The likelihood of a driver's decision can be estimated through a logistic distribution. 

The choice modeling approach is simple and understandable in predicting the choice of the driver. It assumes the independence of the input variables, enabling practitioners to easily grasp the individual impact of each factor. Moreover, it is based on the probabilistic choice theory \cite{mcfadden1972conditional} which accounts for the probabilistic nature of decision-making and is beneficial for uncertainty estimation. Therefore, this approach is widely used in driver utility analysis. For example, in \cite{Ashkrof2022}, the utility function is employed to estimate the effects of various hypothetical attributes such as age, experience level, education level, and monetary information of the request. Similarly, in \cite{Gao2022}, the utility function serves as a prediction model for estimating the acceptance probability of the driver.

\begin{figure}[!ht]
  \centering
  \includegraphics[width=0.4\textwidth]{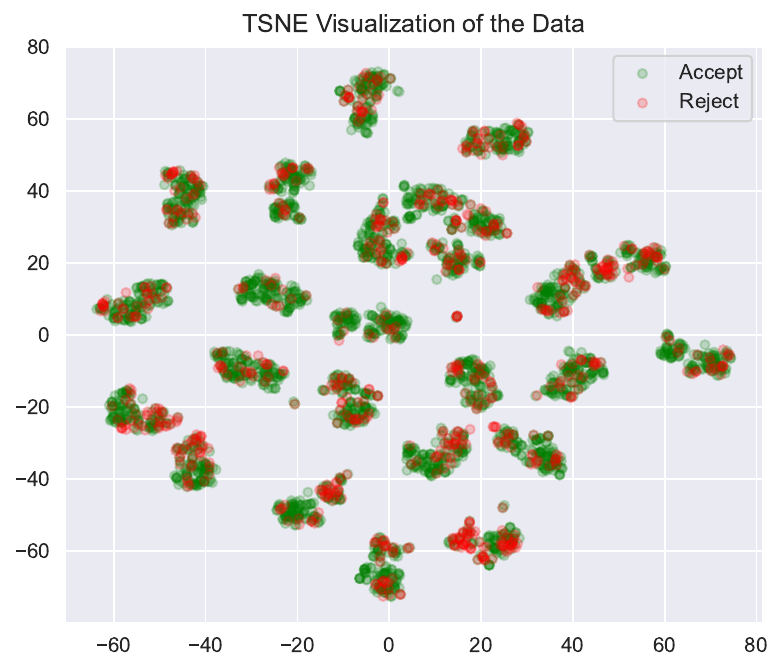}
  \caption{Visualization of the latent distribution of the ride-sourcing data reveals that the data cluster into different groups, and within each group, the acceptance and rejection records are not linearly separable.}
  \label{fig1}
\end{figure}

However, a driver's decision-making process is intricate, the variables may change the preference of the driver, leading to variations in choices. For instance, a driver with a higher expected income might prioritize the fare of a request more than a driver with a lower expected income. The linearity of the utility function could fail to account for such internal nonlinear relationships, potentially leading to poor prediction accuracy. Fig.~\ref{fig1} shows the latent distribution of real-world data gathered from a survey \cite{Ashkrof2022} on driver decision-making. In each cluster (representing data with similar distribution), it is challenging to differentiate between rejected and accepted data. This suggests that drivers may not necessarily make similar decisions when receiving ride requests with similar information due to personal preferences.

One straightforward way to address this problem is to use any nonlinear off-the-shelf machine learning model for the prediction instead of the linear regression. In \cite{wang2021comparing}, Wang et al. evaluate hundreds of sophisticated machine learning classifiers for discrete choice modeling. They reported that among all the tested methods, the ensemble models and deep neural networks achieve the best prediction performance, while the basic choice modeling approaches get a lower accuracy and have a higher computational cost as the number of alternatives increases. 

Although deep learning and ensemble methods are known for their impressive predictive capabilities, they are often criticized for their lack of explainability. In this particular application scenario,  the practitioner may prefer an interpretable model that can elucidate which attributes play key roles in the driver's decision-making process. The understanding of the model prediction can help in designing corresponding monetary strategies to increase the driver's acceptance rate. 

To this end, this paper aims to harness the strengths of deep learning models to develop an explainable predictive approach tailored for analyzing individual driver utility. It can potentially provide insights for designing the matching strategy. Overall, our contribution could be summarized as follows:
\begin{itemize}
    \item We propose a method that leverages hypernetwork architectures and ensemble learning for learning drivers' personalized utility functions. We show that our method can maintain good interpretability, uncertainty estimation and satisfactory prediction accuracy.
    \item We utilize the proposed method to train a utility function and apply it to analyze the general determinants influencing ride acceptance choices, with results aligned with intuitive expectations. We also demonstrate how different attributes can impact an individual's decision using the personalized utility function.
\end{itemize}

\section{Linear Random Utility Function}\label{ruf}
Given the choice set, and a set of variables that affect the choice of an alternative. The utility function can be described as:
\begin{equation}
    U = V + \epsilon.
    \label{utility}
\end{equation}
In the above equation, $V$ and $\epsilon$ represent the systematic contributor and the error term, respectively. The systematic utility is the weighted linear combination of all the attributes: $V = \sum_i w_ix_i$. The Random Utility Maximization (RUM) method is commonly employed to estimate the coefficients of the linear systematic utility. In practice, the coefficients can be calibrated through the data that we use in different scenarios. 

A linear utility function can well represent how each variables contribute to the final decision. However, we posit that the coefficients should be adaptable on the basis of the driver's personal details and the information of the request. One approach that may be considered is to utilize a function that represents the coefficients of the utility function. The function will take as inputs the distinctive information pertaining to both the driver and the ride request. The following section will introduce how we can achieve this goal.

\section{Ensemble Hypernetworks}
The concept of using a deep neural network to generate the hyperparameters of another predictive neural network has been explored in previous studies \cite{DBLP:conf/iclr/HaDL17, chauhan2023brief}. The network that used for hyperparameter generation is called a hypernetwork. In the literature, Kadra et al. \cite{DBLP:journals/corr/abs-2305-13072} are the first to utilize the hypernetwork to generate interpretable linear models. They demonstrated effectiveness of deep hypernetworks for tabular data prediction and image data classification through extensive experiments. Given a dataset $\mathcal{D} = \{(\mathbf{x}_n, y_n)\}$, where $\mathbf{x}_n \in \mathbb{R}^{F}$ is an $F$-dimensional data point and $y_n$ is the corresponding label, the optimization objective function for the parameters of the hypernetwork can be described by the following equations:
\begin{align*}
    \theta^{*} = \arg\min_{\theta\in \Theta} \sum_{n=1}^{N} \Big[ 
    &\mathcal{L}(y_n, f(\hat{y}_n|\mathbf{x}_n;\theta))) \\
    &+ \lambda \left\Vert w(\mathbf{x}_n;\theta)\right\Vert_{1} \\
    &+ (1-\lambda) \left\Vert w(\mathbf{x}_n;\theta)\right\Vert_{2} \Big].
\end{align*}

Where $\theta$ represents the optimal parameters for the hypernetwork, and $w(\cdot):\mathbb{R}^F\rightarrow \mathbb{R}^{F
+1}$ represents the output weights of the features and bias term, \[
f(\hat{y}_n|\mathbf{x}_n;\theta) = \frac{1}{1 + e^{-\left(w(\mathbf{x}_n;\theta)^T\mathbf{x}_n + b\right)}}.
\], The loss function $\mathcal{L}$ in our application domain is set as the Binary Cross Entropy loss (BCE), since the driver utility prediction can be treated as a binary classification problem of accept or reject. The second part of the objective function is an L1-regularized term for maintaining the sparsity of the output weights.

The hypernetwork is an effective tool for generating weights for each attribute in the linear utility function. Thus far, it enables us to obtain personalized weights for each driver. To obtain accurate uncertainty quantification, the ensemble learning techniques \cite{lakshminarayanan2017simple} is used to train multiple hypernetworks to predict the weights in the utility function. The final weights are caculated by averaging the outputs of the hypernetworks.
\begin{equation}
    \Bar{w}(\mathbf{x}_n) = \frac{1}{M}\sum_m^{M} w_m(\mathbf{x}_n;\theta_m).
    \label{avgw}
\end{equation}

\begin{table*}[htbp]
  \centering
  \caption{Models Performance on the driver acceptance behavior dataset. The dataset is split into 5 train-test sets with different random seeds. The results in the table are the average values tested on the train-test sets with the standard deviation.}
    \begin{tabular}{lllllll}
    \hline
    \multicolumn{1}{c}{Models} & \multicolumn{1}{c}{ACC $\left(\uparrow\right)$}   & \multicolumn{1}{c}{AUC $\left(\uparrow\right)$}   & \multicolumn{1}{c}{AUCPR $\left(\uparrow\right)$}  & \multicolumn{1}{c}{ECE $\left(\downarrow\right)$}   & \multicolumn{1}{c}{BS $\left(\downarrow\right)$}    & \multicolumn{1}{c}{NLL $\left(\downarrow\right)$}  \bigstrut\\
    \hline
    Logistic Regression & 0.786 $\pm$ 0.002 &  0.633 $\pm$ 0.017 & 0.334 $\pm$ 0.016 & 0.180 $\pm$ 0.004 & 0.163 $\pm$ 0.002 & 1.447 $\pm$ 0.038 \bigstrut[t]\\
    Decision Tree & 0.764 $\pm$ 0.016 & 0.665 $\pm$ 0.025 & 0.360 $\pm$ 0.020 & 0.117 $\pm$ 0.017 & 0.188 $\pm$ 0.012 & 5.185 $\pm$ 1.086 \\
    Xgboost & 0.794 $\pm$ 0.012 & 0.745 $\pm$ 0.022 & 0.492 $\pm$ 0.020 & 0.100 $\pm$ 0.013 & 0.157 $\pm$ 0.006 & 1.722 $\pm$ 0.152 \\
    TabResNet & 0.779 $\pm$ 0.005 & 0.699 $\pm$ 0.024 & 0.434 $\pm$ 0.030 & 0.139 $\pm$ 0.021 & 0.158 $\pm$ 0.005 & 1.459 $\pm$ 0.082 \\
    HyperNetwork (M=1) & 0.794 $\pm$ 0.004 & 0.704 $\pm$ 0.016 & 0.444 $\pm$ 0.030 & 0.045 $\pm$ 0.007 &  0.155 $\pm$ 0.003 & 1.316 $\pm$ 0.116 \\
    Ens-Hyper (M=5) & 0.802 $\pm$ 0.006 & 0.720 $\pm$ 0.008 & 0.463 $\pm$ 0.012 & 0.034 $\pm$ 0.010 &  0.149 $\pm$ 0.001 & 1.269 $\pm$ 0.059 \bigstrut[b]\\
    \hline
    \end{tabular}%
  \label{tab1}%
\end{table*}%

In the equation above, $\theta_i$ represents the parameters of the $m$-th hypernetwork in the ensemble. Previous studies \cite{lakshminarayanan2017simple, DBLP:conf/iclr/HavasiJFLSLDT21, gal2016dropout} have shown that employing the ensemble models can significantly enhance both the robustness and accuracy of the uncertainty quantification. Then, the predicted probability from one hypernetwork can be obtained using the sigmoid function:
\begin{equation}
    f_m(y|\mathbf{x};\theta_m) = \frac{1}{1 + e^{-(w(\mathbf{x};\theta_m)^T\mathbf{x}+b(\mathbf{x};\theta_m))}}.
\end{equation}
Here, $b(\mathbf{x};\theta_m)$ represents the predicted bias for the utility function, which mimics the error term in \eqref{utility}. The final predicted probability is the average of the predicted probability of each sub-hypernetwork: $f(y|\mathbf{x}) = M^{-1}\sum_{m}^{M}f_m(y|\mathbf{x};\theta_m)$.

\begin{figure}[!ht]
  \centering
  \includegraphics[width=0.45\textwidth]{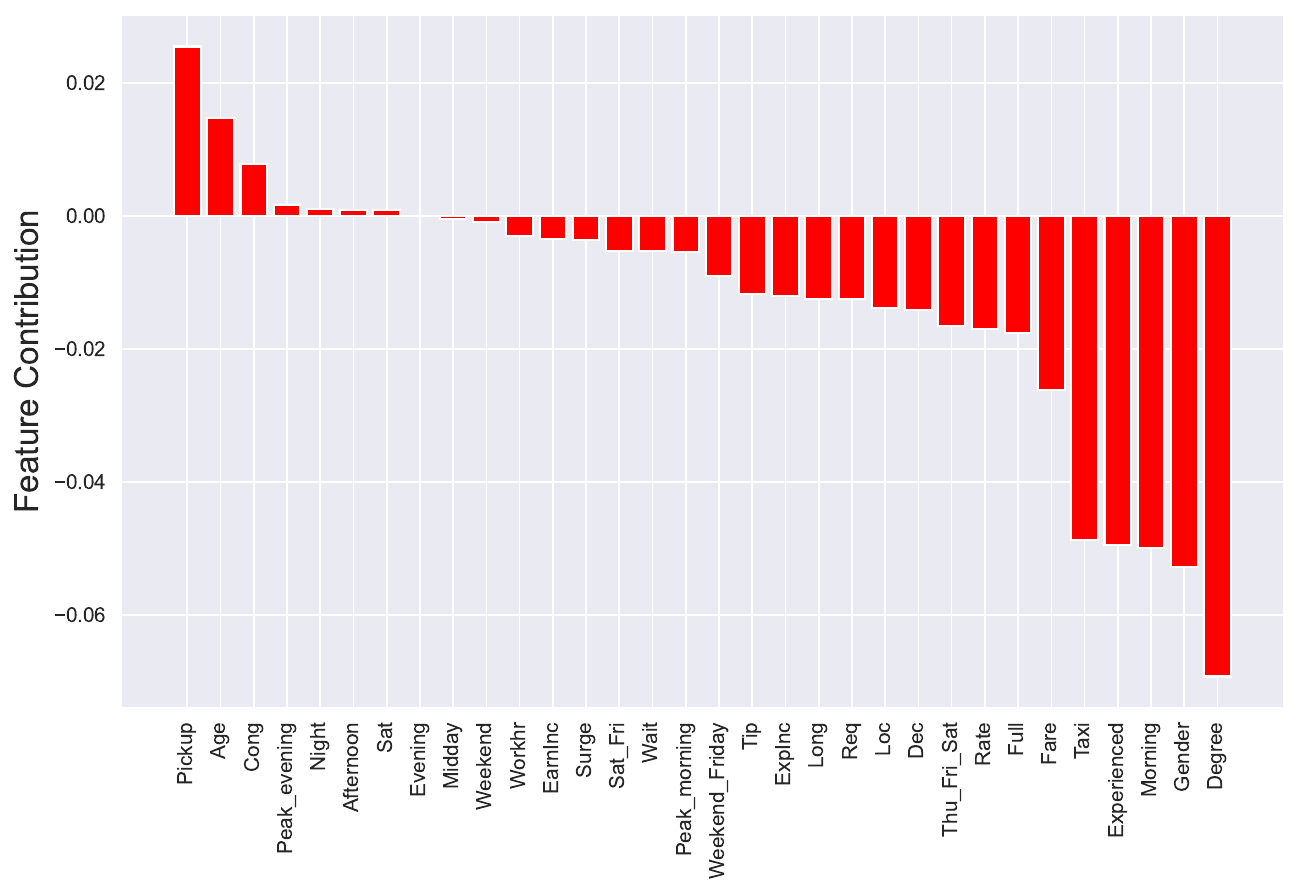}
  \caption{The predicted weights of each feature. The weights are calculated based on \eqref{avgw}.}
  \label{feature_importance}
\end{figure}

\begin{figure*}[t!]
\hspace{-1mm}
\subfloat[]{
\begin{minipage}[t]{0.3\textwidth}
\centering
\includegraphics[width=2in]{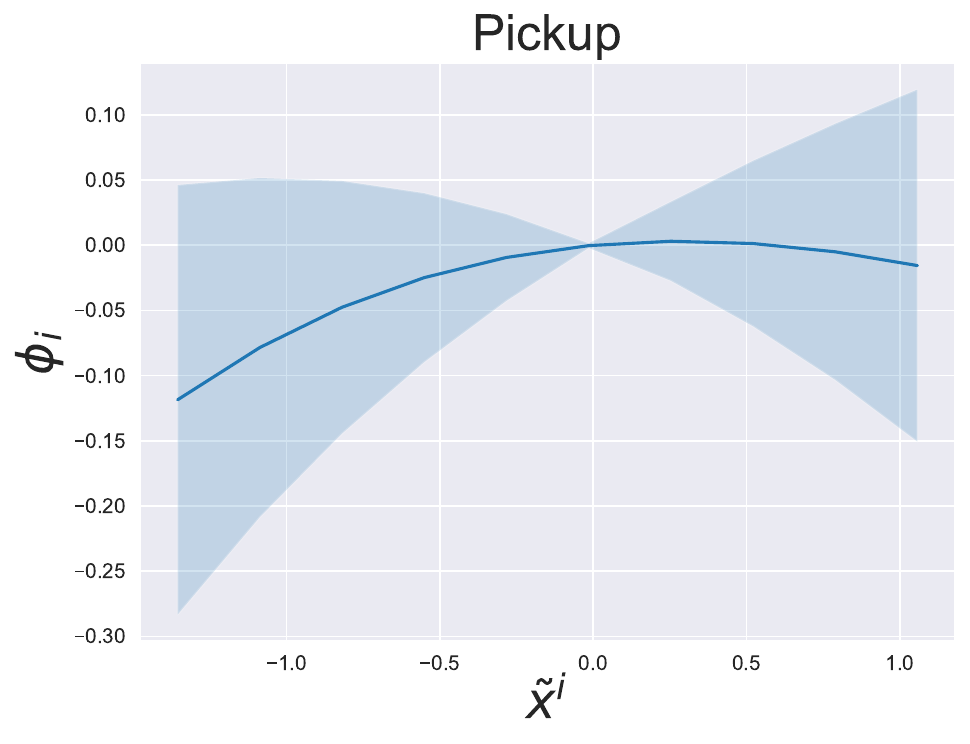}
\label{a}
\end{minipage}%
} \hspace{-1mm}
\subfloat[]{

\begin{minipage}[t]{0.3\textwidth}
\centering
\includegraphics[width=2in]{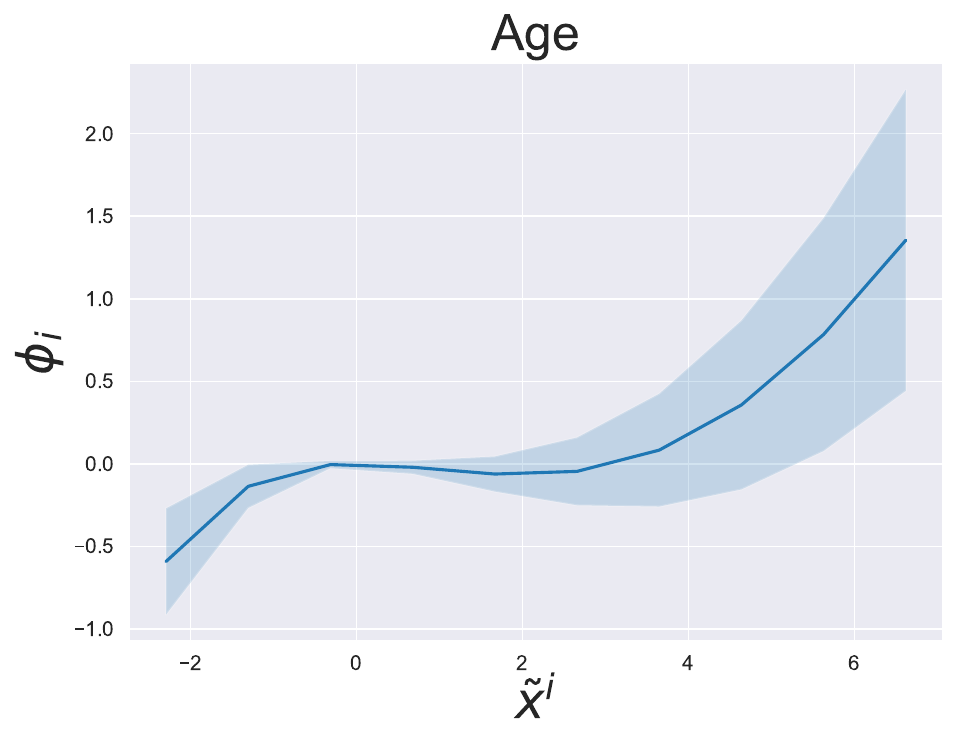}
\label{b}
\end{minipage}%
}\hspace{-1mm}
\subfloat[]{
\begin{minipage}[t]{0.3\textwidth}
\centering
\includegraphics[width=2in]{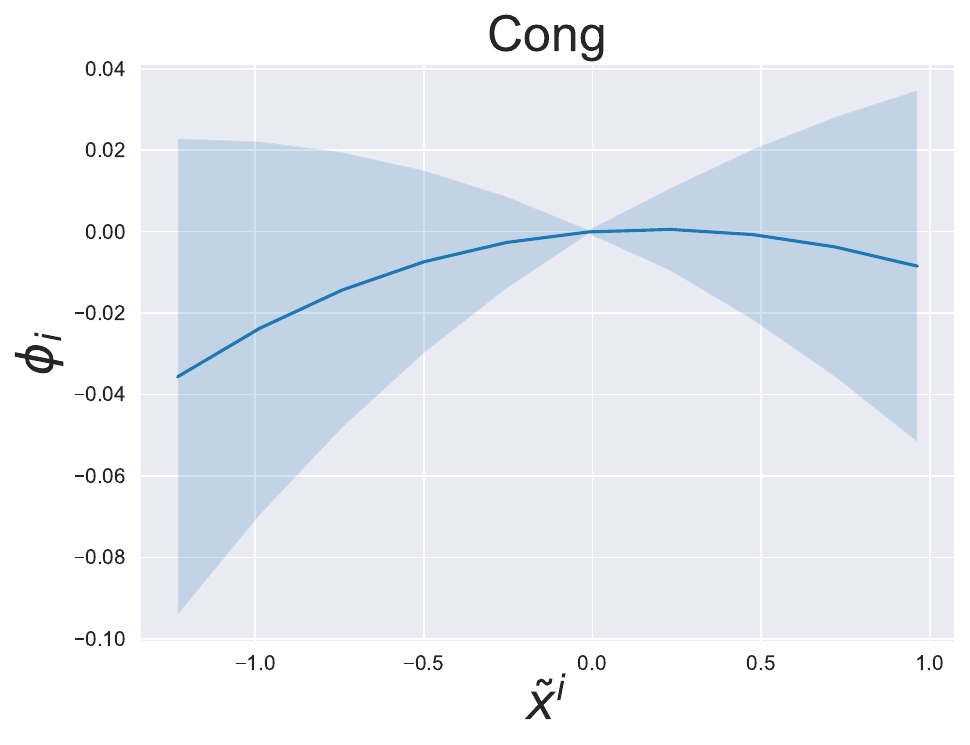}
\label{c}
\end{minipage}%
}
\caption{The feature contribution of $Pickup$ (pick up time), $Age$ (age of the driver) and $Cong$ (delay time due to congestion). The shawdow area represents the standard deviation of $\phi_i$.}
\label{general}
\end{figure*}

\section{Dataset and Model Performance}
In this section, we are going to explore the prediction performance in terms of accuracy and uncertainty estimation. The driver's ride acceptance behavior dataset we use is collected through the cross-sectional stated preference (SP) survey developed by Ashkrof et al. \cite{Ashkrof2022} in 2021. The experimental design incorporates a wide range of attributes, including existing features e.g. pickup time, surge pricing, rider rating, idle time, and hypothetical ones such as traffic congestion, trip fare, and guaranteed tip. In our setting, the monetary features about a request are available for the driver to make a decision, since we can use this information to investigate how the fare and pricing strategy can influence individual driver's decision. Additional details on the dataset can be found in the literature.

The original dataset contains 46 features. We used SHAP (SHapley Additive exPlannations) to calculate feature importance and filtered out the least relevant features. After the screening process, the dataset retains 32 features. After that, all the continuous features in the dataset is standardized by the Z-Score normalization.

There is an imbalance in the quantity of data between the rejected and accepted sets. Therefore, in addition to considering the overall prediction accuracy (ACC), we also employed Area Under the Curve (AUC) and Area Under the Precision-Recall Curve (AUCPR) as the other two metrics. The Expected Calibrated Error (ECE), Brier Score (BS), and Negative Log Likelihood (NLL) \cite{gneiting2007strictly} are used to measure the quality of predictive uncertainty.

Table \ref{tab1} presents the overall performance of the models. We investigate two interpretable machine learning models logistic regression and decision tree, along with one commonly used ensemble method, gradient boosting tree. For the deep learning methods, we employ the Tabular ResNet \cite{kadra2021well} as the baseline and the backbone for the ensemble hypernetworks. From the results, we observe that while ensemble hypernetworks may not achieve the highest prediction accuracy, it provides better uncertainty quantification compared to the selected baseline methods. It offers a balanced consideration between explainability, accuracy, and uncertainty estimation. We also tested the performance of a single hypernetwork (with $M=1$) and found it to be inferior to the ensemble models (with $M=5$) in most metrics, except for ECE.
It is worthnoting that the purpose of this experiment is not to provide a comprehensive comparison between all the methods and to develop a state-of-the-art model, but rather to demonstrate the efficacy of the proposed ensemble hypernetworks model in predicting driver acceptance behavior.

\section{Mining Attributes Correlations}
In this section, we intend to answer the following questions using the pre-trained driver utility function.
\begin{itemize}
    \item What are the key factors influencing driver rejection and how do these factors affect it?
    \item Can ensemble hypernetworks provide conterfactual explanation for individual driver?
\end{itemize}

\subsection{Determinants for General Driver Preference}
Fig.~\ref{feature_importance} presents the predicted importance of each feature (weights in the utility function). We can observe that attributes $Pickup$ (pickup time), $Age$ (age of the driver), and $Cong$ (delay time due to congestion) are three key factors that positively influence the driver's rejection. All monetary factors such as $Fare$, $Tip$, and $Surge$ positively influence driver's acceptance, aligning with our general expectation. Additionally, the attribute $Degree$ significantly contributes to driver acceptance, as educated drivers are more likely to accept a request. These findings are consistent with those in \cite{Ashkrof2022}.

We choose the attributes that with a positive influence to the rejection (i.e. $Pickup$, $Age$ and $Cong$) to see how these variables affect the choice of the driver. Specifically, we can quantify the contribution $\phi_i$ of the $i$-th variable by multiplying the feature value and the expectation of the weights on the dataset:

\begin{equation}
    \phi_i = \mathbb{E}_{\Tilde{\mathbf{x}}}\left[\Tilde{x}^{i}\Bar{w}(\Tilde{\mathbf{x}})_i\right],
\end{equation}
where $\Tilde{x}^{i} \in [\Tilde{x}^{i}_{min}, \Tilde{x}^{i}_{max}]$ is the observed value of the $i$-th attribute of a ride request data $\Tilde{\mathbf{x}}$, bounded by $\Tilde{x}^{i}_{{min}}$ and $\Tilde{x}^{i}_{{max}}$. We then adjust the value of $\Tilde{x}^{i}$ from $\Tilde{x}^{i}_{min}$ to $\Tilde{x}^{i}_{max}$ to observe how $\phi_i$ changes.

The results are shown in Fig.~\ref{general}. A positive contribution indicates that the variable has a positive impact on the driver's rejection behavior. Although we expect longer pickup time and delay to increase the chance that the driver would decline, the effect is non-linear and not monotonic as shown in Fig.~\ref{a} and \ref{c}. Interestingly, the contribution of feature values larger than the mean is decreasing, indicating that there are some requests for which even long pickup times and congestion do not lead to rejection. This suggests that other monetary factors incentivize the driver to accept the request. The result in Fig.~\ref{b} also aligns with our expectation that, generally, younger drivers are more likely to accept rider requests, while drivers older than a certain age are more likely to decline the request.

The above analysis demonstrates how attributes affect the driver's decision globally. However, we can observe that the standard deviation of the contribution is generally large, indicating that the impact of the features varies with different driver profiles and details of the request. In the next section, we will introduce how to analyse the attribute importance for individual drivers using the personalized utility function.

\subsection{Personalized Driver Preference Analysis}
One insightful application of the personalized utility function is that we can use it to analyze the preferences of an individual driver and also perform counterfactual analysis. We choose two true positive instances (the model correctly predicts the driver will decline the request) to illustrate the discrepancy between drivers.

\begin{table*}[htbp]
  \centering
  \caption{Partial information of the two ride request data. \textbf{Remark:} $Rate$: the rate of the rider; $Surge$: surge price for the request; $Tip$: guaranteed tip for the request; $Workhr$: working hours of the driver; $EarnInc$: earned income of the driver; $ExpInc$: driver's expected income.}
  \scalebox{0.8}{
    \begin{tabular}{|r|r|r|r|r|r|r|r|r|r|r|}
    \hline
    \multicolumn{1}{|c|}{\textbf{ID}} & \multicolumn{1}{c|}{\textbf{Rate}} & \multicolumn{1}{c|}{\textbf{Pickup (mins)}} & \multicolumn{1}{c|}{\textbf{Surge (price)}} & \multicolumn{1}{c|}{\textbf{Cong (mins)}} & \multicolumn{1}{c|}{\textbf{Tip (price)}} & \multicolumn{1}{c|}{\textbf{Fare (price)}} & \multicolumn{1}{c|}{\textbf{Workhr (hours)}} & \multicolumn{1}{c|}{\textbf{Age}} & \multicolumn{1}{c|}{\textbf{EarnInc (price)}} & \multicolumn{1}{c|}{\textbf{ExpInc (price)}} \bigstrut\\
    \hline
    68    & 4     & 15    & 0     & 30    & 0     & 8     & 35    & 40    & 450   & 250 \bigstrut\\
    \hline
    133   & 4     & 15    & 0     & 30    & 0     & 8     & 25    & 20    & 50    & 50 \bigstrut\\
    \hline
    \end{tabular}%
  \label{drivers}%
  }
\end{table*}%

\begin{figure*}[t!]
    \hspace{-5mm}
    \subfloat[Driver 68]{
    \begin{minipage}[t]{0.5\textwidth}
    \centering
    \includegraphics[width=3.7in]{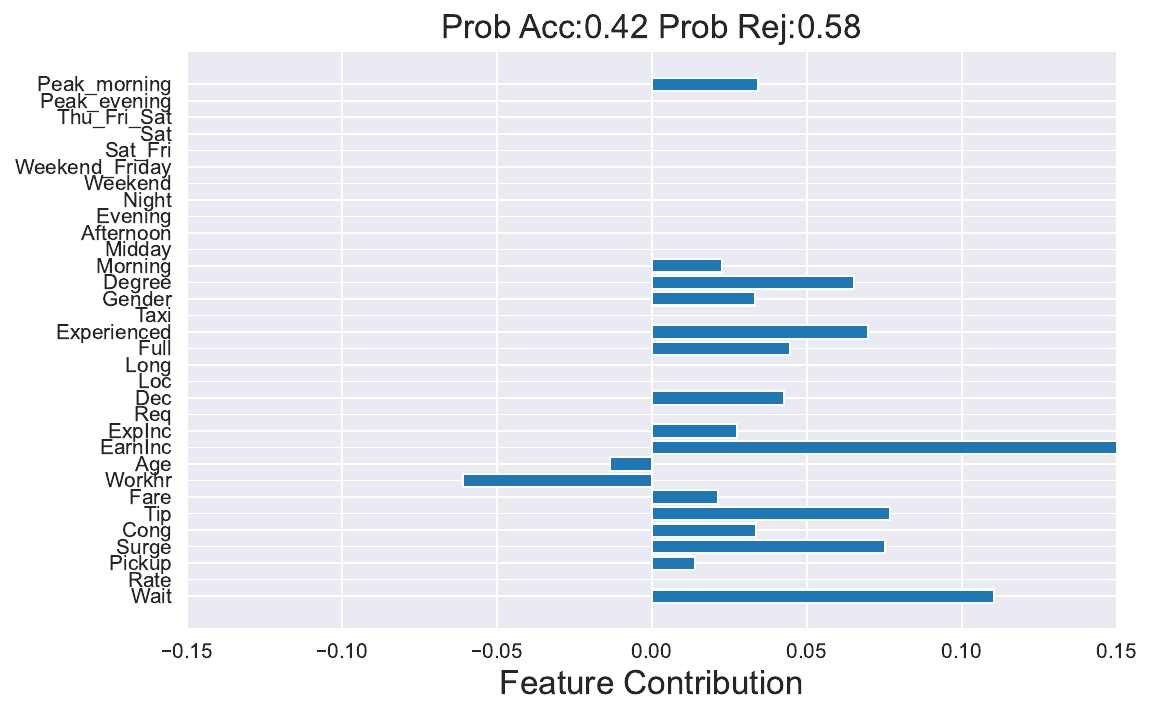}
    \label{driver1}
    \end{minipage}%
    } 
    \hspace{2mm}
    \subfloat[Driver 133]{
    \begin{minipage}[t]{0.5\textwidth}
    \centering
    \includegraphics[width=3.7in]{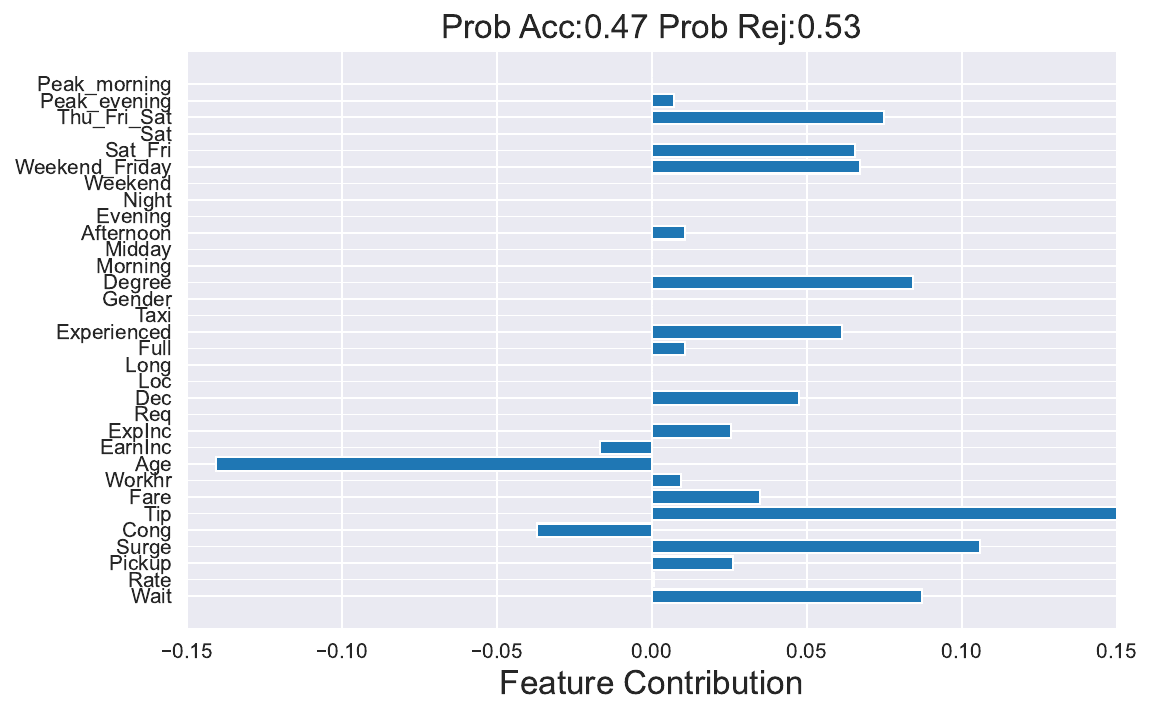}
    \label{driver2}
    \end{minipage}%
    }
    \hspace{-3mm}
    \caption{Personalized feature contribution of two drivers.}
    \label{twodrivers}

\end{figure*}
Partial information of the drivers is presented in Table~\ref{drivers}, and the predicted features contributions of these two drivers are shown in Fig.~\ref{twodrivers}  As we can see in the table, the attribute $Age$ plays a crucial role in the decision of the drivers. The model suggests that the older driver (ID=68), is more likely to accept the request compared to the younger driver (ID=133).

Although both drivers decline the request, different attributes affect the decision differently. For driver 68, the earned income features has a high negative influence on the acceptance, because the driver already earned enough of money and the fare of this request is not high enough for him to accept. Interestingly, although all the monetary variables are the same in these two request data, but their contribution on driver 133's decision is more than driver 68. Interestingly, although all the monetary variables are the same in these two request data, their contribution to driver 133's decision is higher than that of driver 68. This can imply that the younger driver is more sensitive to higher monetary rewards than the older one.

Next, we investigate how the attribute $Fare$ influences the rejection probability of these drivers, i.e., to what extent increasing the fare can incentivize the driver to accept the request. This type of counterfactual analysis can provide insights for designing pricing strategies.

\begin{figure}[ht!]
    \hspace{-5mm}
    \subfloat[Driver 68]{
    \begin{minipage}[t]{0.25\textwidth}
    \centering
    \includegraphics[width=1.7in]{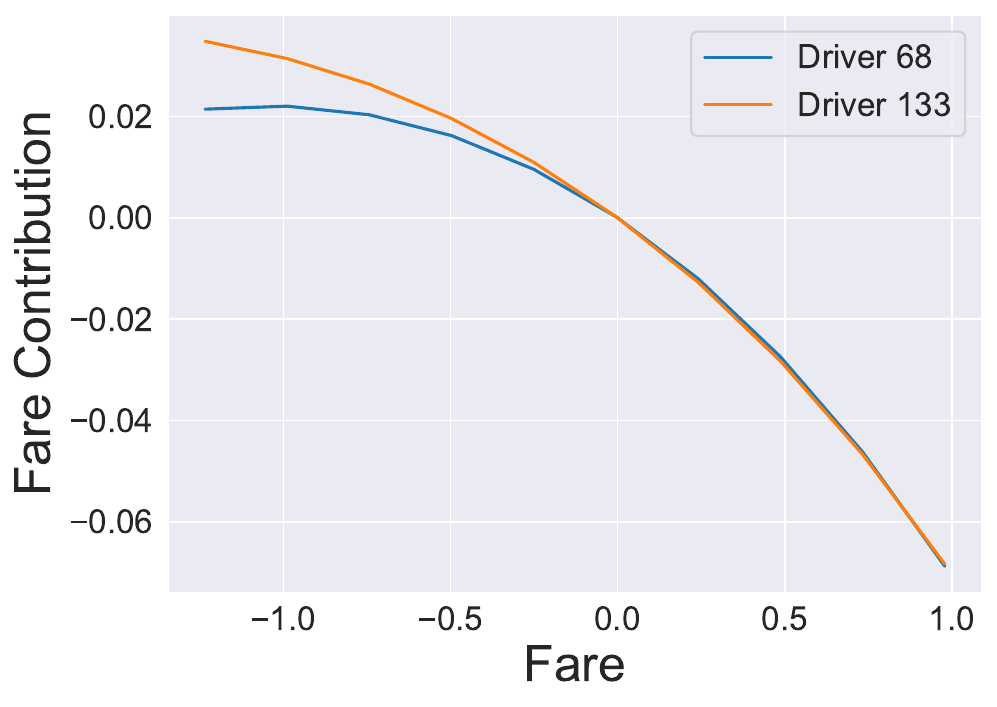}
    \label{fig5a}
    \end{minipage}%
    } 
    \subfloat[Driver 133]{
    \begin{minipage}[t]{0.25\textwidth}
    \centering
    \includegraphics[width=1.7in]{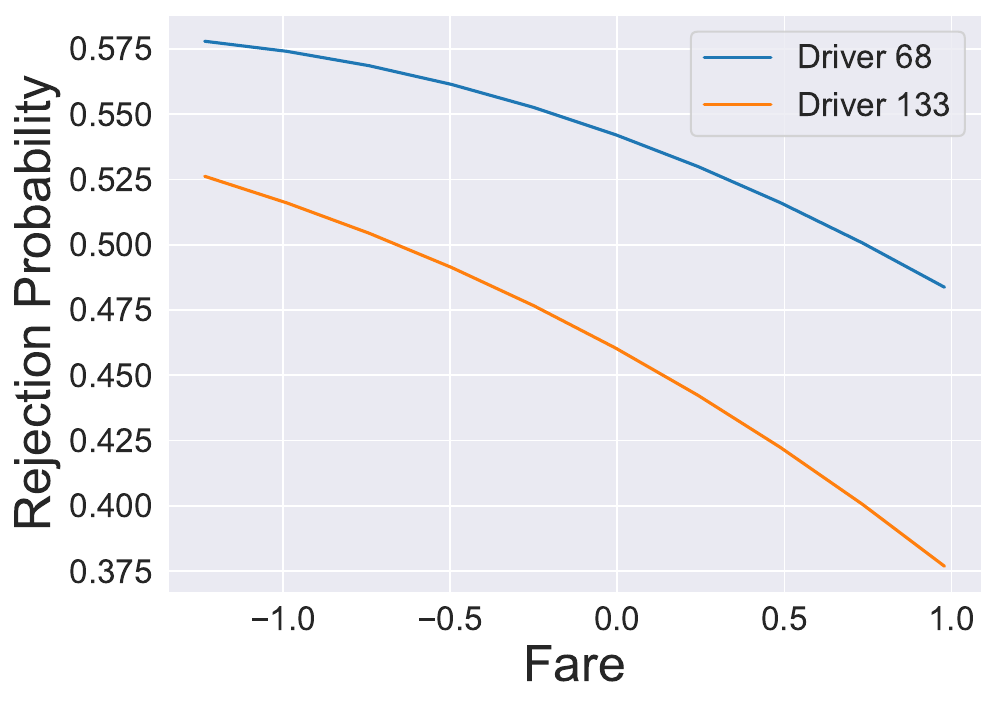}
    \label{fig5b}
    \end{minipage}%
    }
    \hspace{-3mm}

    \caption{(a) The change in fare contribution as the fare changes. (b) The change in rejection probability as the fare changes.}
    \label{fig5}

\end{figure}

Fig.~\ref{fig5a} shows how the fare contribution changes as the fare changes, the result shows that the younger driver (ID=133) is more sensitive to the change of the fare. Intuitively, In Fig.~\ref{fig5b}, we can see that as the fare increases, the rejection probability decreases at different rates. After a certain point, both drivers will accept the request. These results can potentially assist practitioners in designing fares tailored to different drivers.

\section{Conclusion and Future Work}
This paper proposed a method for learning personalized utility functions for drivers in ride-hailing systems. We demonstrates the efficacy of utilizing an ensemble hypernetworks in predicting each driver's utility through individual driver preference analysis. Although similar analyses might be derived from model-agnostic explainers such as SHAP and Local Interpretable Model-Agnostic Explanations (LIME), these techniques fall short in identifying correlations between feature contributions and utility predictions.
Additionally, the personalized utility function derived from this method supports the implementation of dynamic pricing strategies for service operators to incentivize drivers. For instance, the model can evaluate how adjustments in surge pricing might influence a driver's probability of accepting or rejecting assigned orders. This insight helps operators design personalized pricing strategies to optimize the matching rate within the ride-hailing system.

However, the proposed model does have some limitations. Deep learning models are generally depend heavily on the quality of the data. Biases in the dataset, such as imbalanced records related to age or gender, can significantly impair the precision of the model's explanations. This highlights the critical need to collect more comprehensive and higher quality data to optimize the performance of the model.

\section*{Acknowledgment}
We acknowledge the support of TU Delft AI Labs programme. We also thank OpenAI for using their AI tool Chatgpt to improve the writing of this article.

\bibliographystyle{IEEEtran}
\bibliography{main}
\vspace{12pt}

\end{document}